%% file: iclr2023_conference.tex
\documentclass{article} %
\usepackage{MLDD_workshop_2023}
\usepackage{times}
\usepackage{multicol}
\input{math_commands.tex}

\usepackage{hyperref}
\usepackage{url}
\usepackage{algorithm}
\usepackage{algpseudocode}
\usepackage{graphicx}
\usepackage{layouts}
\usepackage{multirow}
\usepackage{caption}
\usepackage{booktabs}
\usepackage[verbose]{wrapfig}
\usepackage{microtype}           %
\usepackage{bookmark}
\usepackage{xspace}
\usepackage{mathtools}

\newcommand{\method}{Subcover\xspace}

\title{The power of motifs as inductive bias\\ for learning molecular distributions}

\author{Johanna Sommer$^*$\textsuperscript{\ensuremath{1,3}}, Leon Hetzel\thanks{equal contribution}\textsuperscript{$\text{\:\:\,}$\ensuremath{1,2,3}}, David Lüdke\textsuperscript{\ensuremath{1}}, Fabian Theis\textsuperscript{\ensuremath{1,2,3}},\\ \textbf{Stephan Günnemann}\textsuperscript{\ensuremath{1,3}} \\[0.8em]
\textsuperscript{\ensuremath{1}}{School of Computation, Information and Technology, Technical University of Munich}\\
\textsuperscript{\ensuremath{2}}{Helmholtz Center for Computational Health, Munich}\\
\textsuperscript{\ensuremath{3}}{Munich Data Science Institute, Technical University of Munich}\\
\texttt{\{jm.sommer, l.hetzel, d.luedke, f.theis, s.guennemann\}@tum.de} \\
}

\begin{document}

\maketitle

\begin{abstract}
Machine learning for molecules holds great potential for efficiently exploring the vast chemical space and thus streamlining the drug discovery process by facilitating the design of new therapeutic molecules.
Deep generative models have shown promising results for molecule generation, but the benefits of specific inductive biases for learning distributions over small graphs are unclear. Our study aims to investigate the impact of subgraph structures and vocabulary design on distribution learning, using small drug molecules as a case study.
To this end, we introduce \method, a new subgraph-based fragmentation scheme, and evaluate it through a two-step variational auto-encoder.
Our results show that \method's improved identification of chemically meaningful subgraphs leads to a relative improvement of the FCD score by 30\%, outperforming previous methods.
Our findings highlight the potential of \method to enhance the performance and scalability of existing methods, contributing to the advancement of drug discovery.

\end{abstract}

\input{chapters/1_introduction}

\input{chapters/3_method}

\input{chapters/4_experiments}
\input{chapters/5_conclusion}

\section*{Acknowledgements and Disclosure of Funding}
JS, LH, and DL are thankful for valuable feedback from Filippo Guerranti, Tobias Schmidt, the DAML group and Theis Lab. LH is supported by the Helmholtz Association under the joint research school “Munich School for Data Science - MUDS”. FJT acknowledges support from the Helmholtz Association’s Initiative and Networking Fund through Helmholtz AI (ZT-I-PF-5-01). FJT further acknowledges support by the BMBF (01IS18053A). In addition, FJT consults for Immunai Inc., Singularity Bio B.V., CytoReason Ltd, and Omniscope Ltd and has an ownership interest in Dermagnostix GmbH and Cellarity. This work was supported by the German Federal Ministry of Education and Research (BMBF) under Grant No. 01IS18036B.

\bibliography{zotero_mod}
\bibliographystyle{iclr2023_conference}

\newpage
\appendix
\input{chapters/appendix.tex}

\end{document}

%% file: math_commands.tex
\usepackage{amsmath,amsfonts,bm}
\usepackage{dsfont}
\newcommand{\atoms}{\ensuremath{\mathcal{A}}}

\newcommand{\bonds}{\ensuremath{\mathbf{B}}}

\newcommand{\vocabsize}{\ensuremath{k}}
\newcommand{\vocab}{\ensuremath{\mathcal{V}}}
\newcommand{\mset}[1]{\ensuremath{\mathcal{M}_{#1}}}
\newcommand{\sset}[1]{\ensuremath{\mathcal{S}_{#1}}}
\newcommand{\fset}[1]{\ensuremath{\mathcal{F}_{#1}}}
\newcommand{\motif}[1]{\ensuremath{m_{#1}}}
\newcommand{\singleatom}[1]{\ensuremath{s_{#1}}}
\newcommand{\fragment}[1]{\ensuremath{f_{#1}}}
\newcommand{\latent}[1]{\ensuremath{\mathbf{z}_{#1}}}
\newcommand{\graph}[1]{\ensuremath{G_{#1}}}
\newcommand{\nodefeatures}[1]{\ensuremath{\mathbf{X}_{#1}}}
\newcommand{\edgefeatures}[1]{\ensuremath{\mathbf{E}_{#1}}}
\newcommand{\fingerprint}[1]{\ensuremath{\mathbf{f}_{#1}}}
\newcommand{\prob}{\ensuremath{\mathds{P}}}
\newcommand{\probG}{\ensuremath{\prob(\mathcal{\graph{}})}}
\newcommand{\card}[1]{\ensuremath{\lvert{#1}\lvert}}

\def\eqref#1{equation~\ref{#1}}

\def\1{\bm{1}}

\DeclareMathAlphabet{\mathsfit}{\encodingdefault}{\sfdefault}{m}{sl}
\SetMathAlphabet{\mathsfit}{bold}{\encodingdefault}{\sfdefault}{bx}{n}

\newcommand{\KL}{D_{\mathrm{KL}}}

%% file: chapters/1_introduction.tex
\section{Introduction}

Generative models for molecules offer a way to create new compounds with specific properties, which can be useful in various fields, including drug discovery, material science, and chemistry \citep{bian_generative_2021, choudhary_recent_2022, hetzel2022predicting, zhu_survey_2022, du_molgensurvey_2022}. The ability to navigate and explore the vast chemical space more efficiently and generate novel molecules in an automated fashion can save time and resources compared to traditional laboratory methods \citep{reymond_enumeration_2012, polishchuk_estimation_2013}. This way, generative models can help discover molecules with unique properties that may not have been found otherwise. Additionally, generative models can assist in optimising existing structures and provide candidate compounds with improved properties \citep{gao_sample_2022}.

To be useful for such applications, any model must be able to abstract molecules in a way that enables it to generate new structures representative of the underlying distribution. A common approach is to learn a continuous latent distribution that captures the discrete structure of molecules \citep{jin_junction_2018, liu_constrained_2018, kusner_grammar_2017}. 
This involves learning the present (i) patterns, such as rings and functional groups, (ii) relationships, such as particular bond types, and (iii) structures, the complex combinations of such geometries, and using that knowledge to generate new, diverse, and yet meaningful molecular structures.

\begin{figure}
\centering
\includegraphics[width=0.9\textwidth]{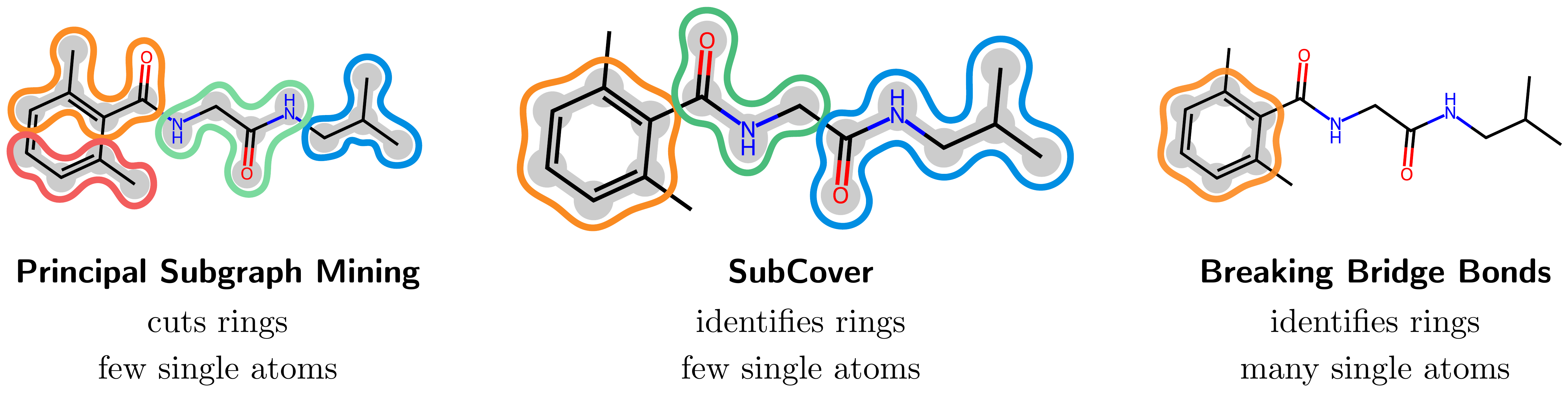}
\caption{Example of a molecule's decomposition according to different fragmentation schemes. Subcover captures cyclic structures while using only few single atoms.} 
\label{fig:herofig}
\end{figure}

There are two predominant approaches for learning distributions of molecules and decoding latent representations back to molecular graphs: atom-based and motif-based approaches \citep{yang_molecule_2022}.
Atom-based models utilise individual atoms as the building blocks of the molecular graph, allowing them, in principle, to model any molecular structure and create highly diverse compounds. However, these models struggle to generate complex and highly symmetric patterns, such as rings \citep{yang_molecule_2022}. Motif-based models, on the other hand, extend the available building blocks with a vocabulary of common fragments or ``motifs'', such as a carbon ring. While a fragment can refer to any molecular substructure, motifs should represent complex geometries that affect molecular properties and provide a chemically informed inductive bias for learning molecular distributions, see \autoref{fig:herofig}.

Yet, the expressive power of motif-based models is determined by the size and quality of the motif vocabulary, as well as the particular decomposition of molecules into motifs. We refer to the combination of the two, vocabulary and decomposition, as a fragmentation scheme. \cite{jin_junction_2018} introduce a scheme that decomposes molecules exclusively into motifs, resulting in a vocabulary that increases with the dataset size and requires the inclusion of many similar motifs. Other approaches overcome the limitation of dataset-dependent vocabulary size by including single atoms in their decomposition \citep{maziarz_learning_2022, kong_molecule_2022b}. These fragmentation schemes, however, fail to provide chemically distinct motifs and low numbers of single atoms simultaneously. Note that the latter is crucial, as a decomposition with too many single atoms hinders learning the molecules' structure. We believe that a well-designed fragmentation scheme provides an inductive bias that balances the ease of learning the molecules' decompositions and their structural properties. As a consequence, learning molecular distributions consists of two main challenges: Identifying the building blocks of an individual compound, which are determined by the respective fragmentation scheme, and learning the structure between those.

In this study, we propose \method, a new approach for fragmenting molecules, and examine its benefits for learning distributions of small molecules. To this end, we rely on a two-step variational auto-encoder (VAE), that extends the work by \citet{kong_molecule_2022b}, as this model type provides good insights into the impact of the fragmentation schemes' inductive biases.

Our main contributions can be summarised as follows:
\begin{itemize}
    \item We investigate the impact of fragmentation methods on learning molecular graph distributions.
    \item We introduce \method, which consistently identifies large, chemically relevant motifs within a molecular graph.
    \item We show that \method improves the learning of molecular graph distributions.
\end{itemize}

%% file: chapters/3_method.tex
\section{Inductive biases through motif vocabularies}
Inductive biases are essential for distribution learning in high-dimensional discrete spaces as they help to reduce their combinatorial complexity. In the case of molecules, identifying motifs---frequently occurring patterns across a molecule dataset---has proven to be a strong prior facilitating the learning of graph distributions. In the following, we introduce the formalities required for building motif vocabularies. Based on this, we present two representative fragmentation schemes, Principled Subgraph Mining (PSM) \citep{kong_molecule_2022b} and Breaking Bridge Bonds (BBB) \citep{maziarz_learning_2022}, and highlight their benefits and drawbacks. On this basis, we introduce a new fragmentation scheme: \method\!.

\paragraph{Preliminaries}
A molecular graph \graph{} can be described by the tuple $\{ \atoms, \bonds \}$, where \atoms\ is the multiset of atoms of size $a = \card{\atoms}$ and $\bonds \in \{0, 1\}^{a \times a \times 3}$ the bonds tensor, indicating both the existence and type of a bond: single-, double-, or triple-bond. Atoms and larger subgraphs can occur multiple times within one molecule and are consequently represented by a multiset.
The goal of graph generation is to learn the distribution \probG\ over the set of graphs $\mathcal{\graph{}} = \{\graph{i}\}$ and subsequently sample new graphs $\graph{\text{new}} \sim \probG$~\citep{zhu_survey_2022}. 

For any given molecule $i$, fragmentation schemes partition the multiset of atoms \atoms$_i$ and their connectivity structure \bonds\ into a multiset of fragments \fset{i} whose union of elements $f$ corresponds again to \atoms, $\atoms=\bigcup_{\fragment{} \in \fset{}} f$. Here, a fragment \fragment{} can be any substructure of \graph{i}, i.e. a single atom \singleatom{} or the elements of a connected subgraph that includes more than one atom. Together with a motif vocabulary \vocab, such a decomposition \fset{} can further be represented as $\fset{}=\mset{} \cup \sset{}$, where \mset{} and \sset{} correspond to the multisets of motifs \motif{}---a motif \motif{} has to include more than one atom---and single atoms \singleatom{}, respectively. Note that we refer to those fragments \fragment{} that are contained in the vocabulary \vocab\ as motifs $\motif{} \coloneqq \fragment{}\in\vocab$. Further, the size of the vocab $\card{\vocab} = \vocabsize$ is often variable, and the decision about which fragments \fragment{i} to include is usually based on a heuristic, for example, the frequency of \fragment{i} in the dataset. Some fragmentation schemes achieve decompositions of \atoms$_{i}$ that are completely described by motifs, $\sset{i}=\emptyset$ \citep{jin_junction_2018}, while others use a combination of the two, $\mset{i}\neq\emptyset$ and $\sset{i}\neq \emptyset$. We associate a connectivity structure with each motif \motif{}, which allows to decompose the bond tensor \bonds\ of a molecule \graph{i} into $\bonds=\bonds_{\mset{i}}\cup\bonds_{\overline{\mathcal{M}}_i}$, where $\bonds_{\mset{i}}$ defines all intra-motif bonds and $\bonds_{\overline{\mathcal{M}}_i}$ all other bonds between motifs and single atoms.

\begin{figure}
\centering
\includegraphics[width=0.95\textwidth]{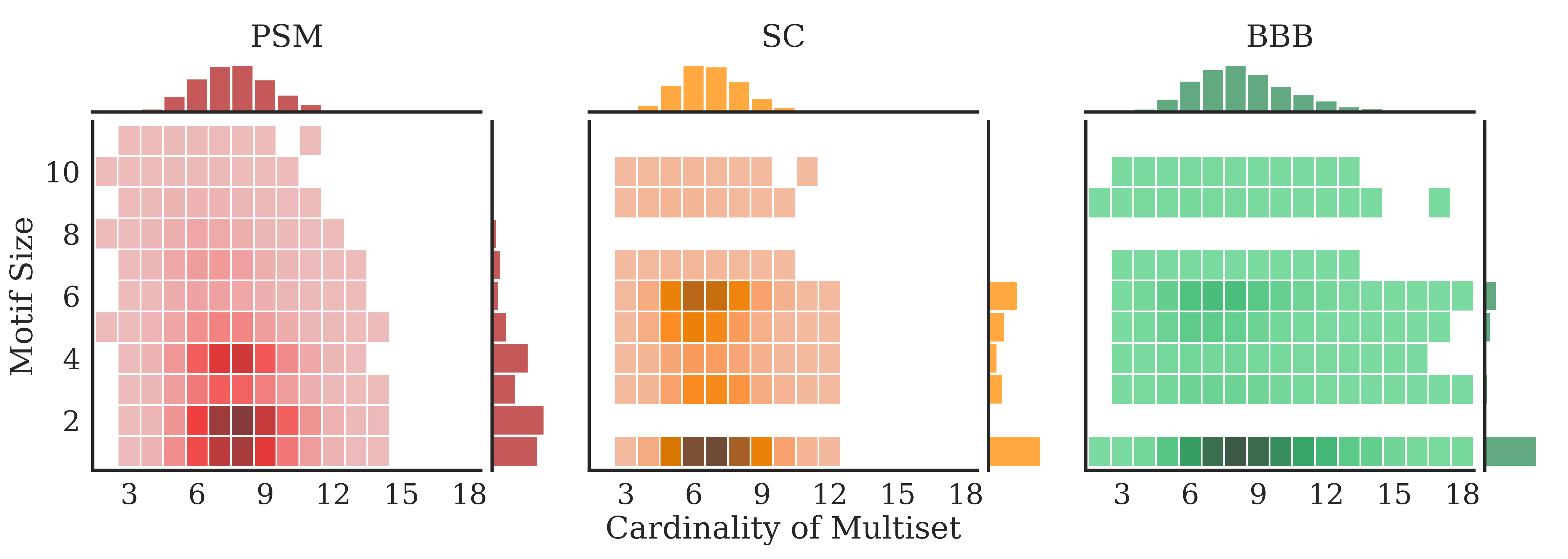}
\caption{Different fragmentation schemes lead to different decomposition sizes $\card{\fset{}}$. BBB leads to larger decompositions due to the increased number of single atoms \sset{}, while PSM mostly decomposes into small motifs. \method\ combines the strength of the two: large motifs and the lowest cardinality among fragmentation schemes.} 
\label{fig:seqlen}
\end{figure}

\paragraph{}\hspace{-13pt}
On the spectrum of fragmentation schemes, we choose two representative approaches to investigate the inductive bias they provide. While \citet{kong_molecule_2022b} identify motifs in a bottom-up manner by starting from single atoms and building up larger structures through merging, the approach by \citet{maziarz_learning_2022} is top-down and leaves all cyclic structures intact. Both methods decompose molecules \graph{} into single atoms \sset{} and motifs \mset{}.

\paragraph{Principled Subgraph Mining} 
\citet{kong_molecule_2022b} initialise the vocabulary generation process from the multiset of single atoms. Using these as initial fragments, the vocabulary \vocab\ is built up one motif at a time.
During each generation step, the current vocabulary is used to decompose all molecules into fragments. The resulting fragments are combined with their neighbours to form new candidate motifs and, among those, the most frequent one is added as a motif to the vocabulary. This way, the vocabulary \vocab\ also includes small motifs, such as a CC chain, which reduces the number of single atoms \sset{i} within a molecule's fragmentation.
To identify the \mset{} and \sset{} multisets of a molecule given a fixed motif vocabulary, the same steps of iterative merging of adjacent fragments must be performed. Starting from single atoms, two fragments are merged only if the resulting fragment is part of the vocabulary \vocab\ and its frequency is the highest among all merged fragments. This process is repeated until no merged fragment is included in the vocabulary anymore.

\paragraph{Breaking Bridge Bonds}
Unlike Principled Subgraph Mining, which is a data-driven approach, the idea of breaking bridge bonds \citep{jin_hierarchical_2020,maziarz_learning_2022} relies on chemical knowledge. During vocabulary construction, a molecule is fragmented by breaking all acyclic bonds adjacent to cyclic structures. For example, the connecting bond between two ring structures will be cut, and likewise, any dangling atom attached to a ring. The final set of \vocabsize\ motifs is selected according to their frequency after applying this fragmentation to all molecules in the dataset. Following \citet{maziarz_learning_2022}, only motifs with a size of at least three atoms are considered. %
Once the vocabulary \vocab\ is fixed, the fragmentation procedure is applied to a graph \graph{i} and those fragments included in \vocab\ are represented as motifs \mset{i}. The remaining fragments are represented as single atoms \sset{i}.
\begin{figure}[t]
    \centering
    \includegraphics[width=0.95\textwidth]{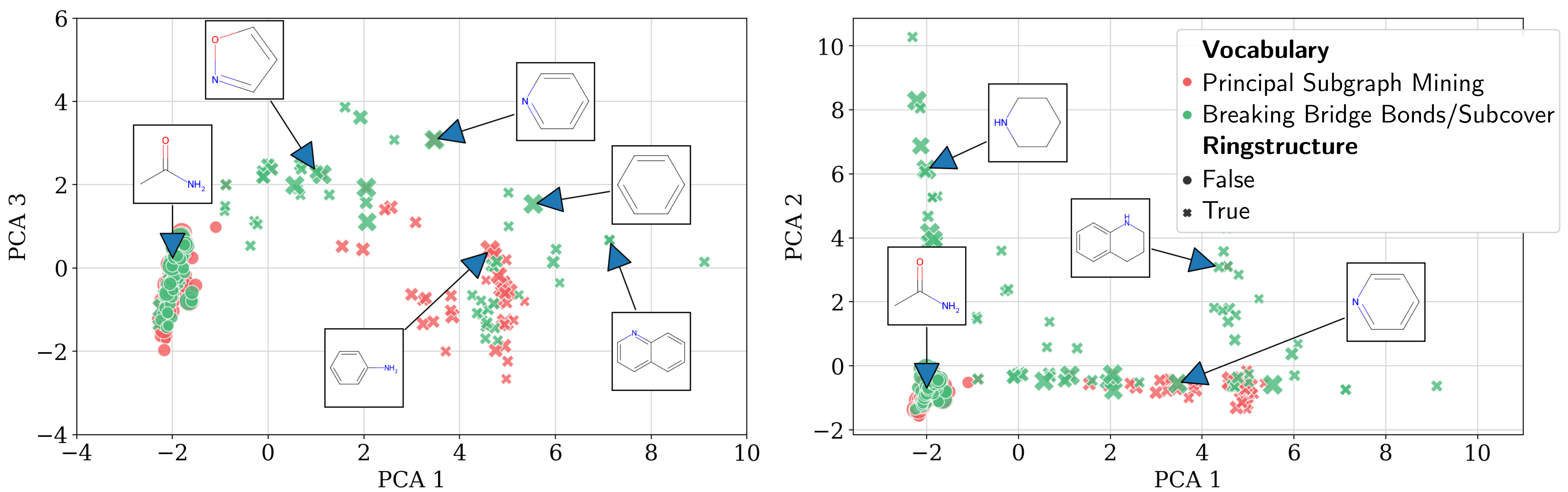}
    \caption{First three principal components for motif fingerprints for PSM and BBB / Subcover vocabularies of size $\vocabsize=128$. PSM identifies fewer rings but many chain-like structures instead. These show little variety in the first three principle components. The size of the markers indicates the frequency of a particular motif, for the BBB / Subcover vocabulary we show the counts of Subcover. In total, the vocabulary of PSM consists of $36\%$ ring-based motifs, while BBB / Subcover has $55\%$ of ring-like structures.}
    \label{fig:motif_fp}
\end{figure}

Both methods, PSM and BBB, reduce the number of single atoms \sset{} through the construction of $\vocab$, which by itself provides a good inductive bias as it decreases the degrees of freedom required to represent molecules. Yet, simply reducing \card{\sset{}} is insufficient. While the bottom-up PSM approach 
\begin{wrapfigure}[28]{r}{0.53\textwidth}
\begin{minipage}[c]{0.5\textwidth}
    \vspace{-2em}
    \begin{algorithm}[H]
    \caption{\method}\label{alg:subcover}
    \begin{algorithmic}
        \Require Molecule $\graph{}$, vocabulary $\vocab$
        \State $\{\fragment{i}\}_{i=0}^{j} \gets \text{BreakBridgeBonds}(\graph{})$ 
        \State $\fset{} \gets \{\}$ 
        \For {$i \in \{0, \dots, j\}$}
        \If{$\fragment{i} \in \vocab$}  
             \State $\fset{} \gets \fset{} ~\cup \fragment{i}$  
        \Else %
             \State $\fset{} \gets \fset{} ~\cup $ FindMInF($\fragment{i}$, $\{\}$)
        \EndIf
        \EndFor
        \Procedure{FindMInF}{$\fragment{}, \fset{}$}
        \State $m^* \gets $ None  
          \For {$\motif{}$ in $\vocab$}
          \If{$\motif{} \subset \fragment{}$ and $\card{\motif{}} \geq\card{\motif{*}}$}
          \If{ and $c(\motif{}) > c(\motif{*})$}
            \State $m^* \gets \motif{}$
          \EndIf
          \EndIf
          \EndFor
          \If{$m^*$ is not None}
          \State $\fragment{} \gets $ DeleteMotifFromFrag($\fragment{}, m^*$)
          \State $\fset{} \gets \fset{} ~\cup m^*$
          \State $\fset{} \gets$ FindMInF($\fragment{}$, $\fset{}$)
          \EndIf
          \State \Return $\fset{}$
        \EndProcedure
    \end{algorithmic}
    \end{algorithm}
\end{minipage}
\end{wrapfigure} 
leads to many molecule decompositions that are completely described by motifs, $\graph{i}=\mset{i}$ and $\sset{i}=\emptyset$, the identified motifs often contain few atoms, cf. \autoref{fig:seqlen}.
In addition, \autoref{fig:motif_fp} shows the qualitative difference between identified motifs revealing an additional drawback of PSM: The motifs in a PSM vocabulary are often chain-like and chemically indistinct. This implies that rings are cut and defining properties of \graph{i} are not reflected by its motif set \mset{i}. In contrast, the top-down BBB approach focuses on leaving cyclic structures intact and considers everything else, i.e. chain-like structures, as ``remainder'' fragments. While this maintains chemical integrity, it also leads to many fragments $\fragment{}\notin\vocab$ being represented as single atoms. This happens even though a motif $\motif{}\in\vocab$ exists which is contained in \fragment{}, $\motif{}\subset\fragment{}$.

\paragraph{\method} 
To combine the strengths of both approaches, we use the same fragmentation as BBB to construct the motif vocabulary \vocab. For the decomposition, however, we go beyond the initial comparison of fragments \fragment{} and motifs $\motif{}\in \vocab$. If $\fragment{}\notin\vocab$, we recursively search for matching subgraphs $\motif{} \subset \fragment{}$. By identifying substructures, we aim to reduce the number of single atoms as PSM does. Should multiple motifs \motif{j} be contained in the fragment \fragment{}, which is often the case, the largest and most frequent motif $\motif{*}$ is selected and the fragment further decomposed, $\fragment{}=\motif{*}\cup \hat{\fragment{}}$. This process is repeated with $\hat{\fragment{}}$ until no more motifs can be identified and the resulting decomposition $\fragment{}=\mset{f}\cup\sset{f}$ is added to the multisets \mset{i} and \sset{i} of the molecule \graph{i}. This reduces the number of single atoms \sset{i} significantly and achieves a more concise, yet chemically meaningful fragmentation, see \autoref{fig:seqlen}. Note that our way of subgraph identification in \method allows mapping structures that are identical up to a charge. While we are not explicitly decoding ions with our model yet, this facilitates the capturing of the molecule's geometry and topology, even for small vocabularies.
A formalisation of \method is presented in Algorithm\,\ref{alg:subcover}.

\section{Two-step VAE}
Fully autoregressive approaches are common for learning distributions of molecules. Yet, the learnt distribution can be difficult to interpret and evaluate independently of the applied corrections and postprocessing. We introduce a two-step VAE approach, which first predicts the multiset of atoms and motifs in a recurrent fashion and subsequently infers their bonds. We build on top of the architecture of \citep{kong_molecule_2022b} and equip our model with multiplicity and motif features to fully harness the provided vocabulary. Our generative model follows a variational encoder-decoder architecture \citep{kingma_auto-encoding_2022}. In doing so, we train an encoder to map the structural graph $\graph{i}$ of a molecule $i$ to a $d$-dimensional latent representation $\latent{i}$. In a subsequent step, the molecule is reconstructed by applying a two-step decoder, that separates the reconstruction of molecule fragments and structure.

\paragraph{Encoder}
The encoder learns the variational posterior $q_\theta(\latent{i}\mid \graph{i})$ that defines the latent representation of each molecule $i$ through $\latent{i}| \graph{i}\sim \mathcal{N}\big(\mathbf{\mu}_\theta(\graph{i}), \mathbf{\sigma}^2_\theta(\graph{i})I\big)$. We parameterise the mean and variance of the approximate posterior as:
\begin{equation}
    \big (\mathbf{\mu}_\theta(\graph{i}), \mathbf{\sigma}^2_\theta(\graph{i})\big ) = h_\theta\big( [g_\theta(\nodefeatures{i}, \edgefeatures{i}), \fingerprint{i}]\big),
\end{equation}
where $g_\theta(\cdot, \cdot)$ is a multi-layer graph neural network (GNN), whose message-passing operation acts on the original molecular graph with the node and edge features $\nodefeatures{i}$ and $\edgefeatures{i}$, respectively, and returns a graph representation. 
The initial node and edge features are learned.
Each node $v \in V$ in $\graph{i}$ is represented by a node feature vector $\mathbf{x}_{v;i} \in \nodefeatures{i}$, that is given by a concatenation of learned embeddings of the corresponding motif and atom identifiers and multiplicity information, as well as its motif fingerprint information.
The connectivity of the molecular graph is encoded by a learned edge embedding that indicates the bond type. For $g_\theta(\cdot, \cdot)$, we leverage a multi-layer transformer convolution backbone \citep{shi_masked_2021} and attain the graph representation of each molecule by a learned aggregation of the transformed node features, see \ref{appendix: Encoder GNN} for more details.
This graph representation is concatenated with a learned embedding of a molecule's fingerprint information $\fingerprint{i}$ before it is mapped to the mean and variance by a multilayer perceptron (MLP) $h_\theta$.

\paragraph{Decoder}
The decoder is trained to reconstruct the molecule given the latent representation $\latent{i}$ sampled from the approximate posterior $q_\theta$.
In a first step, we autoregressively predict the multiset of a molecule's fragments.
We leverage a simple one-layer recurrent neural network (RNN) for the multiset prediction by casting it as a fragment classification task. Formally, the RNN is trained to minimise the negative log-likelihood:
\begin{equation}
    \mathcal{L}_M = - \sum^N_{j=1} \log \prob_{\phi}(\fragment{j}\mid \fragment{j-1}, \dots, \fragment{0}, \mathbf{z}),
\end{equation}
where $\prob_{\phi}(f_j| f_{j-1}, \cdots, \fragment{0}, \mathbf{z})$ is trained to decode the sequence of molecule fragments $\fragment{j}$ starting from a start token $\fragment{0}$ and ending with a stop token.

In a second step, we use the multiset of molecule fragments \fset{i} and latent representation $\latent{i}$ to infer the structure of the molecule, i.e. the existence of bonds and their types.
To predict the molecule's structure, we train two MLPs to parameterise $\prob(\mathbf{b}_{u,v} \mid  \mathbf{z}, \mathbf{x}_u, \mathbf{x}_v)$, where the node features are attained by a multi-layer GNN on the graph of atoms, see \ref{appendix: Decoder GNN} for more details. This graph is fully connected except for the known intra-motif connections.
We leverage the node embeddings from the encoder and contextualise them further with $\latent{i}$ and the multiset of molecule fragments, which is only possible due to our canonical representation of molecules.
Finally, we concatenate the node representations across all GNN layers and predict the adjacencies and bond types to minimise the log-likelihood:
\begin{alignat*}{3}
\mathcal{L}_S = -\sum_{\substack{v,u \in V\\ u \neq v}} \log  \prob(\mathbf{b}_{u,v} \mid \mathbf{z}, \mathbf{x}_u, \mathbf{x}_v).
\end{alignat*}
The VAE is trained to minimise the following loss function:
\begin{equation}
    \mathcal{L} = \beta(\lambda_S \mathcal{L}_S + \lambda_M \mathcal{L}_M + \lambda_V \mathcal{L}_V) + (1-\beta) \KL,
\end{equation}
where $\KL$ is the KL-Divergence between the variational posterior $q_\theta$ and a standard normal prior, and $\mathcal{L}_V$ is a valency regularization loss on the generated molecule. 

In addition to incorporating a valency regularisation into the training process, we employ two forms of post-processing during inference. Specifically, we perform a valency correction that enforces adherence to the valid valency of each atom by restricting the number of bonds accordingly. Further correction can be done via a ``Cycle Breaking" procedure, which limits cyclic structures outside of motifs to a size of either five or six \citep{kong_molecule_2022b}. When applying this procedure, we allow no more than two shared atoms between such cycles. The decision on which bonds to include is based on the model's confidence about its prediction.

%% file: chapters/4_experiments.tex
\section{Experiments}

\begin{figure}
\centering
\includegraphics[width=0.95\textwidth]{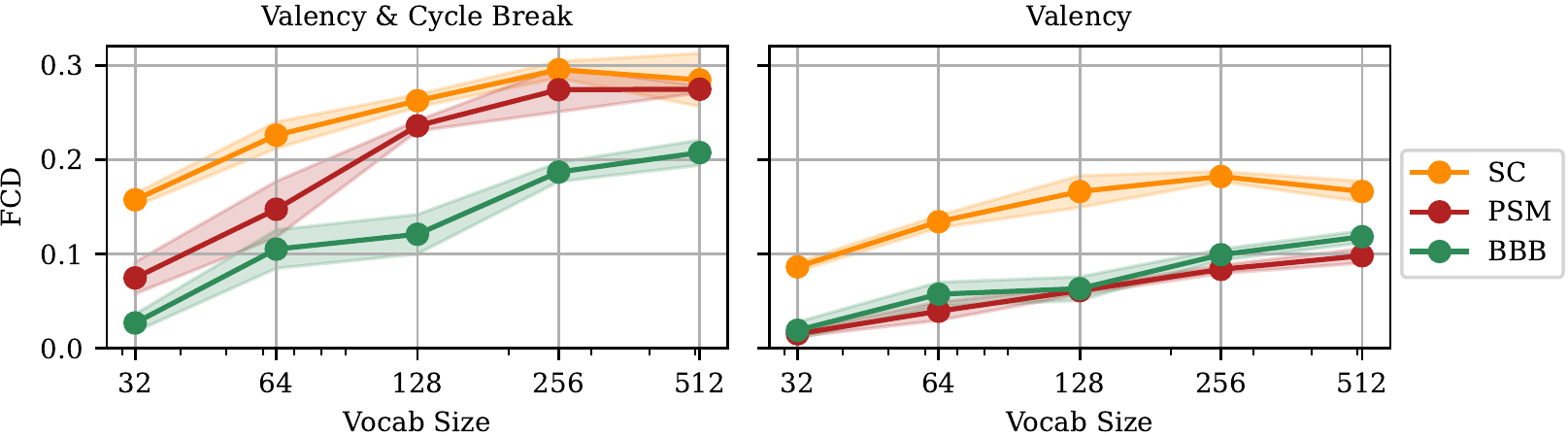}
\caption{FCD achieved with different fragmentation schemes over increasing vocabulary size with cycle breaking postprocessing (left) and only valency correction (right).} 
\label{fig:fcd_over_vocabs}
\end{figure}

\paragraph{Data} We conduct all experiments on the ZINC molecule dataset \citep{kusner_grammar_2017}, which contains a total of 249,457 molecules from the ZINC database \citep{gomez-bombarelli_automatic_2018a}. The dataset is randomly split into training, validation, and test sets with a ratio of 0.8~/ 0.1/ 0.1. All vocabularies are extracted from the entire dataset and the specified vocabulary size does not include single atoms. We train all models for 20 epochs and the hyperparameters used for the baselines are taken from their respective publications. Meanwhile, the hyperparameters for our model are described in \autoref{appendix:hps}. For every model-vocabulary pair, we report the mean and standard deviation over three random model initialisations. Results for sampling new molecules from the learnt distribution are reported for a sample set of 10,000 molecules.

\paragraph{Comparison of Fragmentation Schemes} The Fréchet Chemnet Distance (FCD) \citep{preuer_frechet_2018} is a metric to determine the distance between two sets of molecular graphs. It takes into account both chemical and biological information, as well as the diversity within each set. The comparison of the fragmentation schemes Subcover (SC), Principled Subgraph Mining (PSM), and Breaking Bridge Bonds (BBB), combined with heavy postprocessing (Valency \& Cycle Break), as well as postprocessing only to ensure chemical validity of bonds (Valency), is depicted in  \autoref{fig:fcd_over_vocabs}. To generate new molecules, we sample latent embeddings $\latent{i} \sim \mathcal{N}\big(0, 1)$ and apply our decoder. Experiments beyond sampling from a standard Gaussian distribution can be found in \autoref{appendix:sample-mod}, where Subcover achieves FCD scores over 0.35. Our results show that SC outperforms the other fragmentation methods. Although the motifs found by BBB may have chemical significance, the high rate of single atoms hinders learning of the distribution over the multiset and, consequently, the molecular structure. The fact that PSM achieves a worse FCD score than SC, even though its single atom rate is the lowest across approaches, indicates that only reducing the number of single atoms in a molecule fragmentation is not sufficient. This is further supported by the FCD scores that the fragmentation schemes achieve without major postprocessing. These findings also suggest that due to the indistinctiveness of the PSM motifs, they are difficult to connect correctly. The fact that PSM can only achieve reasonable FCD scores with cycle correction is likely a result of breaking apart rings, as the model must then learn to create rings and does so excessively during inference. This finding is further substantiated by comparing the frequency of rings of different sizes per molecule as reported in \autoref{appendix:rings}. As the vocabulary size increases, both Subcover and PSM stagnate in performance. Overall, the results indicate superior performance of Subcover in the small-vocabulary settings compared to all fragmentation scheme baselines.
\begin{table}[t]
\centering
\caption{Performance metrics for several graph generation models. Best results \textbf{overall} are bold and underlined for \underline{two-step approaches}. Vocabulary sizes are 256, except for JTVAE, which has a fixed vocabulary size of $\sim780$. Results for varying vocabulary size are in \autoref{appendix:baselines}.}
\label{tab:dist_results}
\begin{tabular}{lcccccc}
\toprule
& FCD $\uparrow$ & KL $\uparrow$ & Int. Div $\uparrow$ & SA $\downarrow$ & QED $\uparrow$ \\
\midrule
JT-VAE  & 0.70 $\pm$ 0.006 & 0.95 $\pm$ 0.004 &   0.86 $\pm$ 0.001 & 0.89 $\pm$ 0.006 & 0.07 $\pm$ 0.003 \\
MoLer  & \textbf{0.80 $\pm$ 0.016} & \textbf{0.98 $\pm$ 0.002} &   0.87 $\pm$ 0.001 & 0.56 $\pm$ 0.032 & 0.10 $\pm$ 0.002 \\
\midrule
PS-VAE (CB) & 0.23 $\pm$ 0.001 & \underline{0.83 $\pm$ 0.001} &   \underline{\textbf{0.89 $\pm$ 0.001}} & 1.86 $\pm$ 0.001 & 0.14 $\pm$ 0.001 \\
PS-VAE (V) & 0.08 $\pm$ 0.001 & 0.67 $\pm$ 0.001 &   \underline{\textbf{0.89 $\pm$ 0.001}} & 2.85 $\pm$ 0.010 & \underline{\textbf{0.17 $\pm$ 0.002}} \\
Ours (CB) & \underline{0.30 $\pm$ 0.009} & 0.82 $\pm$ 0.010 &   0.86 $\pm$ 0.002 & \underline{\textbf{0.52 $\pm$ 0.029}} & 0.04 $\pm$ 0.009 \\
Ours (V)  & 0.15 $\pm$ 0.004 & 0.70 $\pm$ 0.007 &   0.88 $\pm$ 0.002 & 1.94 $\pm$ 0.028 & 0.09 $\pm$ 0.007 \\
\bottomrule
\end{tabular}
\end{table}
\paragraph{Molecular Distribution Learning} Finally, to contextualise the results of this study, we provide a comparison of our model with the best-performing fragmentation scheme, Subcover, to three state-of-the-art molecule generation methods: Junction Tree VAE (JTVAE) \citep{jin_junction_2018}, MoLer \cite{maziarz_learning_2022} and PS-VAE \cite{kong_molecule_2022b}. We defer an overview of related work for molecule generation methods to \autoref{appendix:related_work}. As per \citep{brown_guacamol_2019, polykovskiy_molecular_2020}, the Fréchet Chemnet Distance, the Kullback-Leibler Divergence, and Internal Diversity are normalised to a scale of $\left[0,1\right]$, with higher values indicating superior performance. Besides FCD, the KL metric measures the Kullback-Leibler Divergence of two sets of properties with respect to physiochemical properties and the Internal Diversity specifies distances of molecules within the generated set based on Tanimoto similarity \citep{benhenda_chemgan_2017}. Additionally, we report results on the quantitative estimate of drug-likeness (QED) \citep{bickerton_quantifying_2012} as well as the synthetic accessability of drug-like molecules (SA) \citep{ertl_estimation_2009}. Our model trained with \method shows improved results on the FCD metric compared to the two-step PS-VAE approach, especially without correction. All evaluated models perform consistent w.r.t. diversity of the generated molecules. Specifically, both MoLer and our method generate synthetically accessible molecules, emphasising the significance of chemical rules for molecule fragmentation and vocabulary construction. Among the autoregressive models, MoLer demonstrates strong performance across all metrics. The findings of this study regarding the impact of inductive biases are also relevant to autoregressive models. Although they possess a more effective approach to handling individual atoms, they, too, can benefit from a consistent and meaningful fragmentation scheme.

%% file: chapters/5_conclusion.tex
\section{Conclusion}
We present \method, a novel fragmentation scheme that combines the benefits of both data-driven and rule-based techniques to provide an effective inductive bias for learning molecular distributions. 
\method outperforms existing fragmentation methods by producing more meaningful motifs and utilising the vocabulary more efficiently, thus reducing single atoms.
These attributes are shown to be crucial for accurately learning distributions over molecular graphs. Subcover has the potential to improve the performance of various other graph generation models, including autoregressive models. We believe that the findings of this work will not only be useful for the advancement of molecular graph generation but also have broader implications for related fields, such as drug discovery, materials science, and computational biology.

%% file: chapters/appendix.tex
\section{Method Details}
\subsection{Encoder GNN}\label{appendix: Encoder GNN}
Our encoder GNN leverages a transformer convolution backbone and computes the transformed node features at layer $k$ as follows:

\begin{equation}
    \mathbf{x}^{(k+1)}_i = \mathbf{W}_1 \mathbf{x}^{(k)}_i +
                \sum_{j \in \mathcal{N}(i)} \alpha_{i,j}^{(k)} \left(
                \mathbf{W}_2 \mathbf{x}^{(k)}_{j} + \mathbf{W}_6 \mathbf{e}_{ij}
                \right),
\end{equation}

where $\mathbf{W}_n$ are learned matrices and the attention coefficients $\alpha_{i,j}^{(k)}$ are computed via multi-head dot product attention:

\begin{equation}
    \alpha_{i,j}^{(k)} = \textrm{softmax} \left(
                \frac{(\mathbf{W}_3\mathbf{x}^{(k)}_i)^{\top}
                (\mathbf{W}_4\mathbf{x}^{(k)}_j + \mathbf{W}_6 \mathbf{e}_{ij})}
                {\sqrt{h}} \right).
\end{equation}

To attain graph features $g_i \in \mathbb{R}^{(H+1)\cdot d_{enc}}$ we aggregate the transformed node features by applying a sum aggregation and a learned weighted aggregation:

\begin{equation}
    g_i = [\sum_{v\in V_i} \mathbf{x}^{(k)}_v; \sum_{v\in V_i} \textrm{MLP}_{\theta_1}(\mathbf{x}^{(k)}_1) \text{MLP}_{\sigma_1}(\mathbf{x}^{(k)}_1); \dots; \sum_{v\in V_i} \textrm{MLP}_{\theta_H}(\mathbf{x}^{(k)}_v) \textrm{MLP}_{\sigma_H}(\mathbf{x}^{(k)}_v)],
\end{equation}

where $\textrm{MLP}_{\theta_h}$ and $\textrm{MLP}_{\sigma_h}$ are two-layer MLPs of feature head $h$.

\subsection{Decoder GNN}\label{appendix: Decoder GNN}

The decoder GNN is based on an MLP-based edgeweighter for which we employ residual connection between the layers:

\begin{equation}
    \begin{aligned}
        \mathbf{m}_i^{(l+1)} &= \sum_{j \in \mathcal{N}(i)} \textrm{MLP}_{\phi_{l}^{edge}} (\mathbf{z},
        \mathbf{h}_i^{(l)}) \cdot
        \mathbf{\Theta} \cdot \mathbf{h}_j^{(l)}\\
        \mathbf{h}_i^{(l+1)} &= \textrm{MLP}_{\phi_l^{node}}(\mathbf{m}_i^{(l+1)},
        \mathbf{h}_i^{(l)}),
    \end{aligned}
\end{equation}

where $\mathbf{\Theta}$ is a weight matrix, and $\textrm{MLP}_{\phi_{l}^{edge}}$ and $\textrm{MLP}_{\phi_l^{node}}$ are a two- and three-layer MLP, respectively.

\section{Model Hyperparameters}\label{appendix:hps}
\begin{table}[h]
\caption{Hyperparameters for our two-step VAE.}\label{tab:hps}
\centering
\begin{center}
\begin{tabular}{| c | c |}
Optimizer & Adam \\
Learning rate & 0.000874 \\
Learning rate decay & 0.99 \\
Batch size & 64 \\
Gradient clipping magnitude & 3 \\
\midrule
Loss weights $\lambda_S, \lambda_M, \lambda_v$ & 1 \\ 
$\beta$ initalization & 0 \\ 
$\beta$ maximum & 0.2 \\ 
$\beta$ annealing start & 0 \\
$\beta$ annealing frequency & 100 \\
$\beta$ annealing step size & 1.4347e-05  \\
Valency penalty initalization & 0 \\ 
Valency penalty maximum & 0.001 \\ 
Valency penalty annealing start & 0 \\
Valency penalty annealing frequency & 1000 \\
Valency penalty annealing step size & 0.00025 \\
\midrule
Encoder layers & 4 \\
Decoder layers & 1 \\
Latent representation size & 128 \\
Atom identity embedding size & 50 \\
Motif identity embedding size & 75 \\
Atom multiplicity embedding size & 30 \\
Motif multiplicity embedding size & 30 \\
Global graph feature size & 115 \\
Motif feature size & 100 \\
Edge decoder connection weight & 0.3 \\
\end{tabular}
\end{center}
\end{table}

\autoref{tab:hps} details all model and training hyperparameters used to obtain the results for our two-step VAE. These parameters are fixed for training across vocabulary sizes and types.

\section{Baseline Results over varying vocabulary sizes}\label{appendix:baselines}

In addition to \autoref{tab:dist_results}, we expand on these result by reporting the Fréchet Chemnet Distance, the Kullback-Leibler divergence, the Internal Diversity I, the synthetic accessability (SA) as well as the druglikeness (QED) across vocabulary sizes in \autoref{tab:baseline_resu}. For calculation of these metrics we rely on the open-source implementations RDKit \cite{landrum_rdkit_2010}, the GuacaMol benchmark \citep{brown_guacamol_2019} and the MOSES benchmark \citep{polykovskiy_molecular_2020}.

\begin{table}[!htbp]
\caption{Results for learning distributions over molecular graphs across several graph generation models and increasing vocabulary size.}\label{tab:baseline_resu}
\centering
\begin{tabular}{l|c|ccccc}
\toprule
               &   \vocabsize  & FCD & KL & Int. Div. & SA & QED \\
\toprule
JTVAE & 780 & 0.70 $\pm$ 0.01 & 0.95 $\pm$ 0.01 &      0.86 $\pm$ 0.01 & 0.89 $\pm$ 0.01 & 0.07 $\pm$ 0.01 \\
\midrule
Moler & 32  & 0.71 $\pm$ 0.01 & 0.98 $\pm$ 0.01 &      0.87 $\pm$ 0.01 & 0.49 $\pm$ 0.03 & 0.09 $\pm$ 0.00 \\
               & 64  & 0.74 $\pm$ 0.01 & 0.99 $\pm$ 0.01 &      0.87 $\pm$ 0.01 & 0.52 $\pm$ 0.03 & 0.09 $\pm$ 0.01 \\
               & 128 & 0.76 $\pm$ 0.01 & 0.98 $\pm$ 0.01 &      0.87 $\pm$ 0.01 & 0.52 $\pm$ 0.01 & 0.09 $\pm$ 0.01 \\
               & 256 & 0.79 $\pm$ 0.02 & 0.98 $\pm$ 0.01 &      0.87 $\pm$ 0.01 & 0.56 $\pm$ 0.03 & 0.10 $\pm$ 0.01 \\
               & 512 & 0.82 $\pm$ 0.01 & 0.98 $\pm$ 0.01 &      0.87 $\pm$ 0.01 & 0.59 $\pm$ 0.04 & 0.09 $\pm$ 0.01 \\
\midrule
PS-VAE & 32  & 0.09 $\pm$ 0.01 & 0.78 $\pm$ 0.01 &      0.90 $\pm$ 0.01 & 2.34 $\pm$ 0.00 & 0.16 $\pm$ 0.00 \\
               & 64  & 0.14 $\pm$ 0.01 & 0.81 $\pm$ 0.01 &      0.90 $\pm$ 0.01 & 2.08 $\pm$ 0.02 & 0.16 $\pm$ 0.01 \\
               & 128 & 0.18 $\pm$ 0.01 & 0.81 $\pm$ 0.01 &      0.89 $\pm$ 0.01 & 2.02 $\pm$ 0.00 & 0.15 $\pm$ 0.01 \\
               & 256 & 0.23 $\pm$ 0.01 & 0.83 $\pm$ 0.01 &      0.89 $\pm$ 0.01 & 1.88 $\pm$ 0.01 & 0.14 $\pm$ 0.01 \\
               & 512 & 0.30 $\pm$ 0.01 & 0.84 $\pm$ 0.01 &      0.89 $\pm$ 0.01 & 1.74 $\pm$ 0.00 & 0.13 $\pm$ 0.01 \\
\midrule
Ours & 32  & 0.16 $\pm$ 0.01 & 0.73 $\pm$ 0.03 &      0.86 $\pm$ 0.01 & 0.93 $\pm$ 0.07 & 0.05 $\pm$ 0.02 \\
               & 64  & 0.23 $\pm$ 0.01 & 0.79 $\pm$ 0.01 &      0.87 $\pm$ 0.01 & 0.81 $\pm$ 0.02 & 0.04 $\pm$ 0.01 \\
               & 128 & 0.26 $\pm$ 0.01 & 0.82 $\pm$ 0.03 &      0.86 $\pm$ 0.01 & 0.69 $\pm$ 0.04 & 0.04 $\pm$ 0.01 \\
               & 256 & 0.30 $\pm$ 0.01 & 0.82 $\pm$ 0.01 &      0.86 $\pm$ 0.01 & 0.52 $\pm$ 0.03 & 0.04 $\pm$ 0.01 \\
               & 512 & 0.28 $\pm$ 0.03 & 0.83 $\pm$ 0.01 &      0.87 $\pm$ 0.01 & 0.48 $\pm$ 0.01 & 0.04 $\pm$ 0.01 \\
\bottomrule
\end{tabular}

\end{table}

\section{Sampling from the Training Latent Distribution}\label{appendix:sample-mod}
\begin{figure}
\centering
\includegraphics[width=0.85\textwidth]{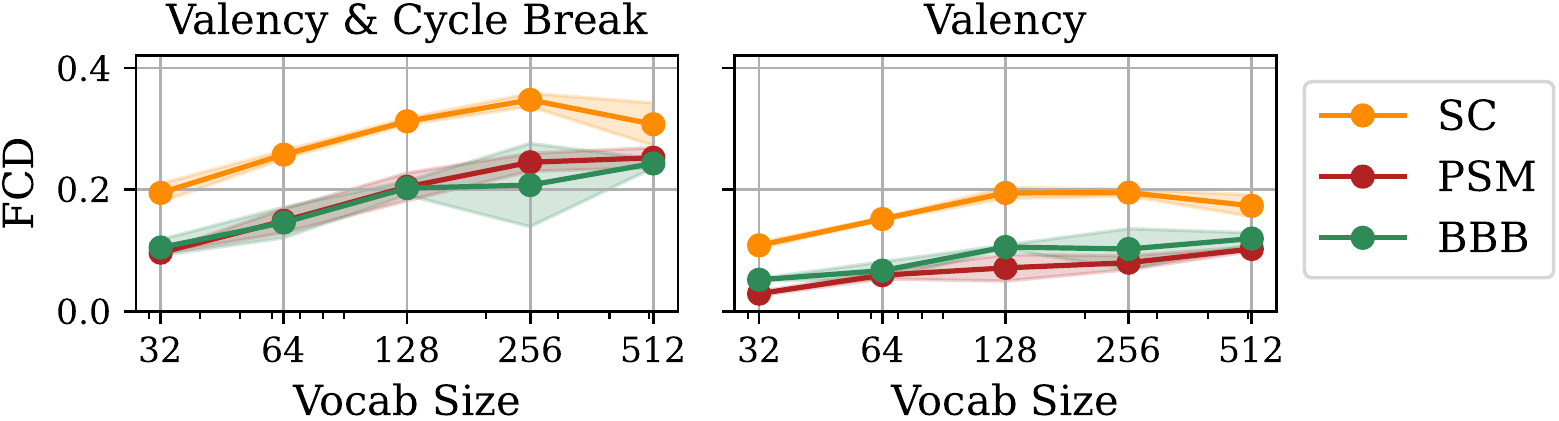}
\caption{FCD achieved with different fragmentation schemes over increasing vocabulary size when sampling from a modified latent distribution.} 
\label{fig:fcd_sampling_mod}
\end{figure}

While a Variational Autoencoder is trained to structure its latent space according to $\latent{i} \sim \mathcal{N}\big(0, 1)$, the latent space may be structured slightly differently even for a fully trained model. Specifically, a VAE faces a tradeoff between reconstruction performance on the training data and the KL Divergence between the approximate posterior and the prior distribution. This tradeoff is controlled via the $\beta$ parameter during training. In a further evaluation, we modify the sampling procedure of our method slightly by first encoding $D=10.000$ molecules, calculating their mean and standard deviation and then sampling new molecules according to: 
\begin{equation}
    \latent{} \sim \big (\frac{1}{D} \sum_{j=0}^D \mathbf{\mu}_\theta(\graph{j}), \frac{1}{D} \sum_{j=0}^D \mathbf{\sigma}^2_\theta(\graph{j})\big )
\end{equation}

The results of this evaluation are shown in \autoref{fig:fcd_sampling_mod}. The findings reveal that with a modified distribution, PSM and BBB exhibit comparable FCD results. The results indicate that PSM does not gain significant improvements in FCD scores as a result of modifying the latent distribution compared to \autoref{fig:fcd_over_vocabs}. BBB on the other hand improves significantly over its performance on a standard Gaussian distribution, suggesting that the latent representations derived from BBB fragmentation make the concise structuring of the latent space more challenging. In this context, Subcover demonstrates FCD scores exceeding 0.35 with a moderately-sized vocabulary.

\section{Frequency of ring structures}\label{appendix:rings}
\begin{figure}
\centering
\includegraphics[width=\textwidth]{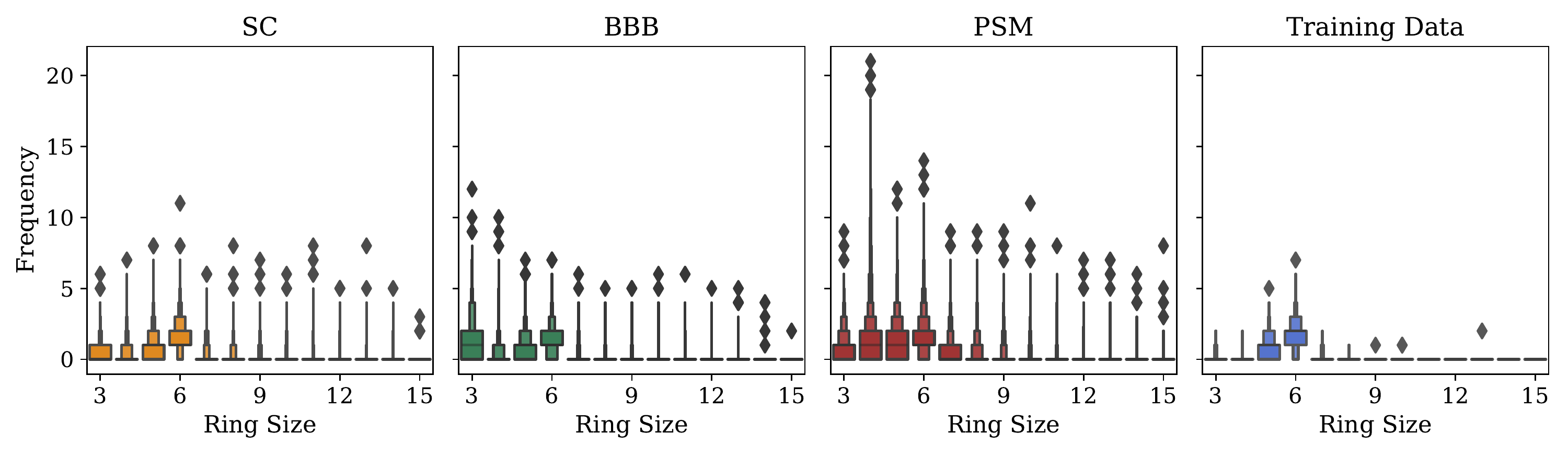}
\caption{Frequency of ring structures with different ring sizes per molecule for samples from SC, PSM, and BBB compared to the frequencies observed in the training data. } 
\label{fig:cycle_count}
\end{figure}
As an additional experiment, we evaluated the frequency of ring structures per sampled molecules from SC, PSM and BBB compared to the training data and report the results in \autoref{fig:cycle_count}.
While all three methods tend to produce more rings per molecule than what was present in the training data, SC shows the closest overall match to the distribution. BBB produces slightly fewer rings for the higher ring sizes but generates more small rings. On the other hand, PSM produces an excessive number of rings per molecule, with some extreme cases of up to 20 rings of size four.

\section{Related Work}\label{appendix:related_work}
Methods for molecule generation differ primarily in the chosen molecular representation. While first attempts characterise molecules through their SMILES or SELFIES strings \citet{gomez-bombarelli_automatic_2018a, nigam_augmenting_2019, kusner_grammar_2017}, focus soon shifted towards a 2D graph representation of molecules due to desirable properties such as permutation invariance. In a parallel line of work, 3D graph generation is concerned with predicting the geometry of a molecule from its 2D representation, called conformer. We refer the reader to \citep{du_molgensurvey_2022} for an overview of 1D, 2D, and 3D generation methods.

\citet{yang_molecule_2022} separate the 2D molecule generation literature into three subcategories: all-at-once, node-based
and fragment-based. Many methods, both node-based \linebreak\citep{khemchandani_deepgraphmolgen_2020, shi_graphaf_2020, popova_molecularrnn_2019, mercado_graph_2021, luo_graphdf_2021, liu_constrained_2018, li_learning_2018, assouel_defactor_2018, ahn_spanning_2021} and fragment-based \citep{yu_molecular_2022, you_graph_2019, yang_hit_2021, lim_scaffold_2020, kajino_molecular_2019, jin_hierarchical_2020, bengio_flow_2021}, rely on the autoregressive decoding of the molecular representation, either by attaching single atoms or larger fragments, such as motifs or scaffolds. In \autoref{tab:baseline_resu}, we compare to MoLer, a molecular graph generation model trained to extend structural scaffolds using the BBB fragmentation scheme \cite{maziarz_learning_2022}. Additionally, we report results for Junction Tree Variational Autoencoder (JTVAE) \citep{jin_junction_2018}, a model that autoregressively builds a junction tree of a molecule and extracts a molecular graph based purely on fragments.

Within the all-at-once category, \citet{simonovsky_graphvae_2018} and \citet{ma_constrained_2018} predict the graph adjacency, as well as the node and edge features in one step. Here, the number of considered nodes usually is fixed manually before inference. \citet{liu_graphebm_2021} follow the same approach, but replace the VAE-based architecture with an energy-based learning technique. Similarly, \citet{decao_molgan_2018} predict the entire molecular graph at once and train within a GAN framework including a permuation invariant discriminator. In contrast, \citet{zang_moflow_2020} first generate a bond tensor through normalizing flows and subsequently assign node features to the graph whose computatiton is conditioned on the same bond tensor. Lastly, \citet{bresson_two_2019} as well as \citet{shepherd_graph_2020} first predict the set of atoms present in the molecular graph, and in a second step compute the connection among these. \citet{samanta_nevae_2019} has a similar generation procedure, while additionally learning the 3D coordinates of the resulting molecular graph. Lastly, \citep{kong_molecule_2022b} first decode the set of nodes and fragments autoregressively via an RNN and then predict attachements of nodes and fragments.